# Plant robots: harnessing growth actuation of plants for locomotion and object manipulation

*Kazuya Murakami, Misao Sato, Momoki Kubota, and Jun Shintake\**

K. Murakami, M. Sato, M. Kubota, Prof. J. Shintake
Shintake Research Group, Department of Mechanical and Intelligent Systems Engineering,
The University of Electro-Communications, 1-5-1 Chofugaoka, Chofu, 182-8585 Tokyo, Japan
*E-mail: shintake@uec.ac.jp

**Abstract**
Plants display physical displacements during their growth due to photosynthesis, which converts light into chemical energy. This can be interpreted as plants acting as actuators with a built-in power source. This paper presents a method to create plant robots that move and perform tasks by harnessing the actuation output of plants: displacement and force generated from the growing process. As the target plant, radish sprouts are employed, and their displacement and force are characterized, followed by the calculation of power and energy densities. Based on the characterization, two different plant robots are designed and fabricated: a rotational robot and a gripper. The former demonstrates ground locomotion, achieving a travel distance of 14.6 mm with an average speed of 0.8 mm/h. The latter demonstrates the picking and placing of an object with a 0.1-g mass by the light-controlled open-close motion of plant fingers. A good agreement between the experimental and model values is observed in the specific data of the mobile robot, suggesting that obtaining the actuation characteristics of plants can enable the design and prediction of behavior in plant robots. These results pave the way for the realization of novel types of environmentally friendly and sustainable robots.

## 1. Introduction

Sustainability and eco-friendliness are crucial functionalities for minimizing environmental impacts and ensuring the responsible use of resources in the rapidly advancing robotics field.[1–3] One approach to realize these functionalities in robots is to incorporate biodegradable materials. In the natural environment, where accidents or malfunctions could compromise retrieval, biodegradable materials prove especially effective for robots. These characteristics make biodegradable robots well-suited for tasks such as exploration, monitoring, transportation, and rescue missions. To achieve these tasks, robots require capabilities such as mobility and grasping, which necessitate actuators as an essential element that generate mechanical outputs.

Researchers have developed various types of biodegradable actuators that work under stimulation from air injection, electricity, temperature, humidity, and light.[4–13] While these actuators function well and are often comparable to those made from synthetic materials, they mostly depend on external drive sources, such as compressors and converters. The incorporation of biodegradability into drive sources requires various components for biodegradation, which are currently under development and have limited performance.[14–16] These have limited performance compared to synthetic materials and require further development. It is crucial to ensure the biodegradability of drive sources for a fully biodegradable robot.

Herein, we suggest employing plants as actuators for robots, utilizing eco-friendly drive sources and photosynthesis, a process that turns light into chemical energy. Plants grow through physical displacement, which serves as an actuation stroke. During growth, they also use force to push aside obstacles such as soil and gravel. Furthermore, plants acquire energy from the natural environment, such as sunlight and soil, and are adept at converting this energy necessary for actuation.[17,18] In other words, plants can be considered actuators with drive-source capabilities.

Plant biology typically focuses on the physical movements connected with plant growth and stimulus responses, with many research done solely on observation and classification[19]. However, quantitative understanding of the movements exhibited by plants is limited, and comprehensive analyses of actuation characteristics such as displacement, force, and speed remain absent. Exceptionally, the displacement of mung bean (*Vigna radiata*) under different loads has been characterized as growth actuation quite recently in a study,[20] yet the extent of the analysis is limited. Consequently, there is a lack of knowledge necessary for designing and constructing plant-based robots. Existing studies on integrating plants into mechanical systems include motor-driven robots and interfaces that utilize the light-sensitive properties of plants as sensors,[21,22] devices that close insectivorous plants via electrical stimulation,[23,24] pneumatic actuators with embedded seeds,[25] and plant-driven actuators for human-robot interaction (this reference also showcases concepts of robots that differ from those developed in our study, such as robot bodies that can extend or separate, shape-changing blocks, and a transmission mechanism using gears, although their performance and design and fabrication methods are not specified).[20] However, the characteristics of plants as actuators remain unclear, limiting their incorporation into the physical movements of robots, especially locomotion, which has not been demonstrated.

This study presents a method to create plant robots. In terms of displacement, force, and speed, we investigated the actuation characteristics of radish sprouts as a model plant. Based on the experimental results, two different types of robots were designed: a ground-based locomotion robot and a gripper. We showed that these plant robots can move and pick and put objects, proving the feasibility of using plant growth to power moving robotic systems as a means of creating entirely biodegradable, environmentally friendly robots.

## 2. Results and discussion
### 2.1. Actuation characteristics of radish sprouts

As the model plant to reveal its actuation characteristics, we used radish sprouts, belonging to the *Brassicaceae* family, due to their linear growth, rapid growth cycle (4–10 days), and hydroponic cultivation, which allow for straightforward implementation into the experimental setup and robots.

A critical aspect of an actuator is its input–output relationship. In the early stages of plant growth, light significantly impacts growth characteristics, thereby altering actuation characteristics. Under low light conditions, plants undergo etiolation, characterized by rapid elongation of the hypocotyls in search of light. Conversely, when exposed to adequate light intensity, they experience photomorphogenesis, marked by slowed hypocotyl elongation and spreading leaves.[26] Since plants can alter their growth form in response to light, they can be considered actuators operating with light as an input. To investigate the actuation characteristics of radish sprouts, we examined their displacement and force characteristics in both illuminated and dark environments (see the Experimental Section for additional details). The distinction between the two environments lies in the presence or absence of light.

First, we assessed the plant's displacement. **Figure 1a,b** shows the growth patterns observed in each environment (see also Supplementary Video S1). Evidently, although the plants exhibited a linear growth pattern, their deformation over time was non-linear (Figure 1c). After 40 h, the displacement reached 46.4 ± 10.9 mm in the dark environment (corresponding to an average speed of 1.2 ± 0.3 mm/h) and 16.3 mm ± 5.1 mm in the illuminated environment (corresponding to an average speed of 0.4 ± 0.1 mm/h). The exponential and bounded nature of these growths can be expressed using a growth mode.[27] Figure 1c shows the measurement results for stem length and growth model fitting. The model can be expressed as follows:

$$L(t) = \frac{K}{1 + \left(\frac{K - L(0)}{L(0)}\right)e^{-rt}} \tag{1}$$

where, $L$ represents the length of the plant at time $t$, $r$ is the intrinsic rate of elongation, and $t$ is the maximum elongation limit. We obtained parameters of $r = 0.0792$, $K = 105$ mm in the dark environment and $r = 0.0500$, $K = 65.1$ mm in the light environment. The growth rate at each length can be calculated by the derivative of Equation 1 with respect to $t$. The maximum speeds were 2.09 mm/h at a length of 52.8 mm in the dark environment and 0.91 mm/h at a length of 36.4 mm in the light environment. These results indicated that the growth and changes in characteristics require time. Therefore, when incorporating them into a robot, compared to general actuators, it is necessary to consider long-term operation.

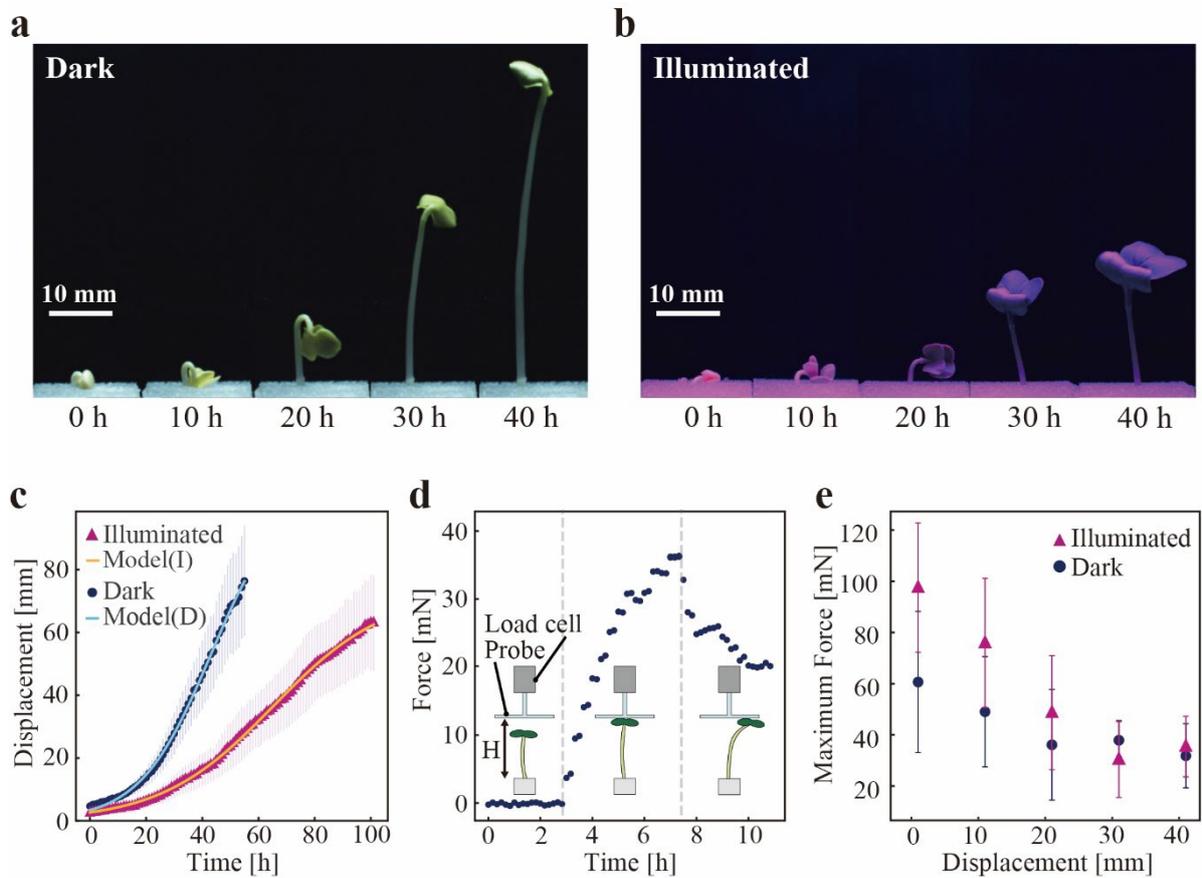

**Figure 1.** Growth behavior, displacement, and force characterization results. (a) Observed growth of the plant in a dark environment. (b) Observed growth of the plant in an illuminated

environment. (c) Measured displacement as a function of the elapsed time. (d) A representative measured data of force as a function of the elapsed time. (e) Measured maximum force at each displacement (on the horizontal axis, 0 represents the point where the seeds complete their germination process).

We then measured the force characteristics of the plant. Figure 1d shows representative measured data of the force as a function of the elapsed time and an illustration of the measurement process (see also Supplementary Video S2). As radish sprouts grew, they contacted the probe of the load cell and pushed gradually along with their growth. When they could no longer withstand the reaction force against the probe, they began to grow horizontally, and the force applied to the load cell started to decrease. Figure 1e plots the measured maximum force at each displacement of the plant. In both environments, the force peaked upon sprouting: $60.5 \pm 27.6$ mN in the dark environment and $97.5 \pm 25.2$ mN in the illuminated environment. The force decreased with increasing displacement. This trend resembled that of conventional actuators, where the force decreases as the stroke increases, implying that plants can function as actuators. The force was greater in the illuminated environment than in the dark environment upon sprouting. However, this difference diminished as the plant grew. This attenuation can be attributed to various factors, including enhanced force from photomorphogenesis and reduced force from growth constraints. Importantly, exposing radish sprouts to light at the initial stage was effective in enhancing their force as actuators. Figure 1e also demonstrates relatively large errors in the measured forces, likely owing to individual differences among the plants—a common issue in plant experiments.[28–33] In the context of this study, the abovementioned individual differences manifest as variations in seed size, sprouting angle, growth rate and direction, tip and stem shape, and contact angle with the load cell probe. To minimize these errors, one approach involves using physical guides to control plant growth. However, these guides may inhibit the natural growth of plants, complicating measurements of true forces. An alternative approach involves using multiple plants as a single actuator. Notably, in this study, a gripper with multiple plants exhibited much lower errors compared to a single plant, as discussed later.

Based on the measured growth speed and force, the power density of the plant upon sprouting was estimated at $181 \times 10^{-6}$ W/kg in the dark environment and $102 \times 10^{-6}$ W/kg in the illuminated environment. Besides, the energy density upon sprouting was 16.6 J/kg in the dark environment and 26.3 J/kg in the illuminated environment. The power densities were lower than those of typical actuators[34,35] because plant actuation occurs over tens of hours. This provides insight into the applicability of plant-based robotics, as they operate on a different time scale than typical robotic systems.

## 2.2. Plant growth-based mobile robot

We developed a rolling robot, as shown in **Figure 2a**, to demonstrate the mobility of a plant-powered robot. The robot consisted of radish sprouts, a structural frame, and hydrogel with a diameter of 25 mm and a mass of 5 g. The robot moved on the ground by converting the growth-induced displacement into rotational motion. Figure 2a illustrates the arrangement of four radish sprouts at 45-degree intervals, each at a distinct growth stage after sprouting, to facilitate continuous rotational movement. The experiment observed that the radish sprouts were all at an early stage of growth, exerting the greatest force during the initial sprouting stage. Figure 2b shows an example of a radish sprout used in the robot, where the longest sprout that contacts the ground determines the robot's initial position. Starting from this position, the growing radish sprouts exerted a force that pushed the ground and rotated the structure against a rolling resistance of approximately 5 mN (see Experimental Section). As depicted in Figure 2c, as the first radish sprout sufficiently rotated the robot, the second one made contact with the ground and began to contribute to the rotational movement. The repetition of this process enabled continuous rotation.

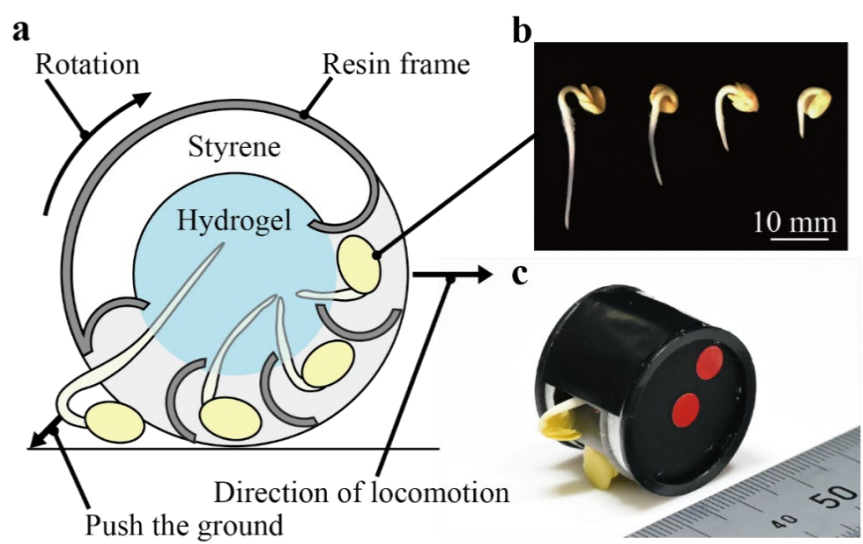

**Figure 2.** (a) Structure of the mobile plant robot. (b) A picture of a set of radish sprouts placed in the robot. (c) The robot rotates and switches to the second radish sprout.

To clarify how the actuation characteristics of plants influence the performance of the robot, we measured its movement in the horizontal plane in both dark and illuminated environments.

**Figure 3a** shows the typical operation of the robot in a dark environment and its internal state (see also Supplementary Video S3). The robot exhibited a rotational motion and subsequent locomotion in the horizontal direction, demonstrating its ability to move based on plant growth. Figure 3b plots the change in horizontal displacement over time and the predicted horizontal displacement based on the growth model. The observed horizontal displacements of the robot were 14.6 mm in the dark environment and 10.4 mm in the illuminated environment. The average speed until 15 h were 0.8 mm/h and 0.7 mm/h for the dark and illuminated environments, respectively. The reason for the shorter travel distance observed in the illuminated environment can be attributed to an increase in water consumption associated with increased transpiration during photosynthesis. Stomata are more likely to open when exposed to light, which facilitates transpiration. This is thought to have caused the robot to rapidly lose water, making it unable to exert the force necessary for rotation. From the start to 10 h, the horizontal displacements in both environments closely matched the model's predicted values for the dark environment. We expected the behavior in the illuminated environment to be similar to the dark environment model because the radish sprouts' contribution to rotation preceded their response to light. To accurately predict the behavior of a robot, a more detailed analysis of actuation characteristics and model design is necessary, considering factors such as light input timing and duration. Nevertheless, the strong agreement between the experimental results and model predictions in the dark environment up to approximately 15 h (with a coefficient of determination of 0.84) indicates that understanding plant actuation characteristics can facilitate the design and behavior prediction of plant robots.

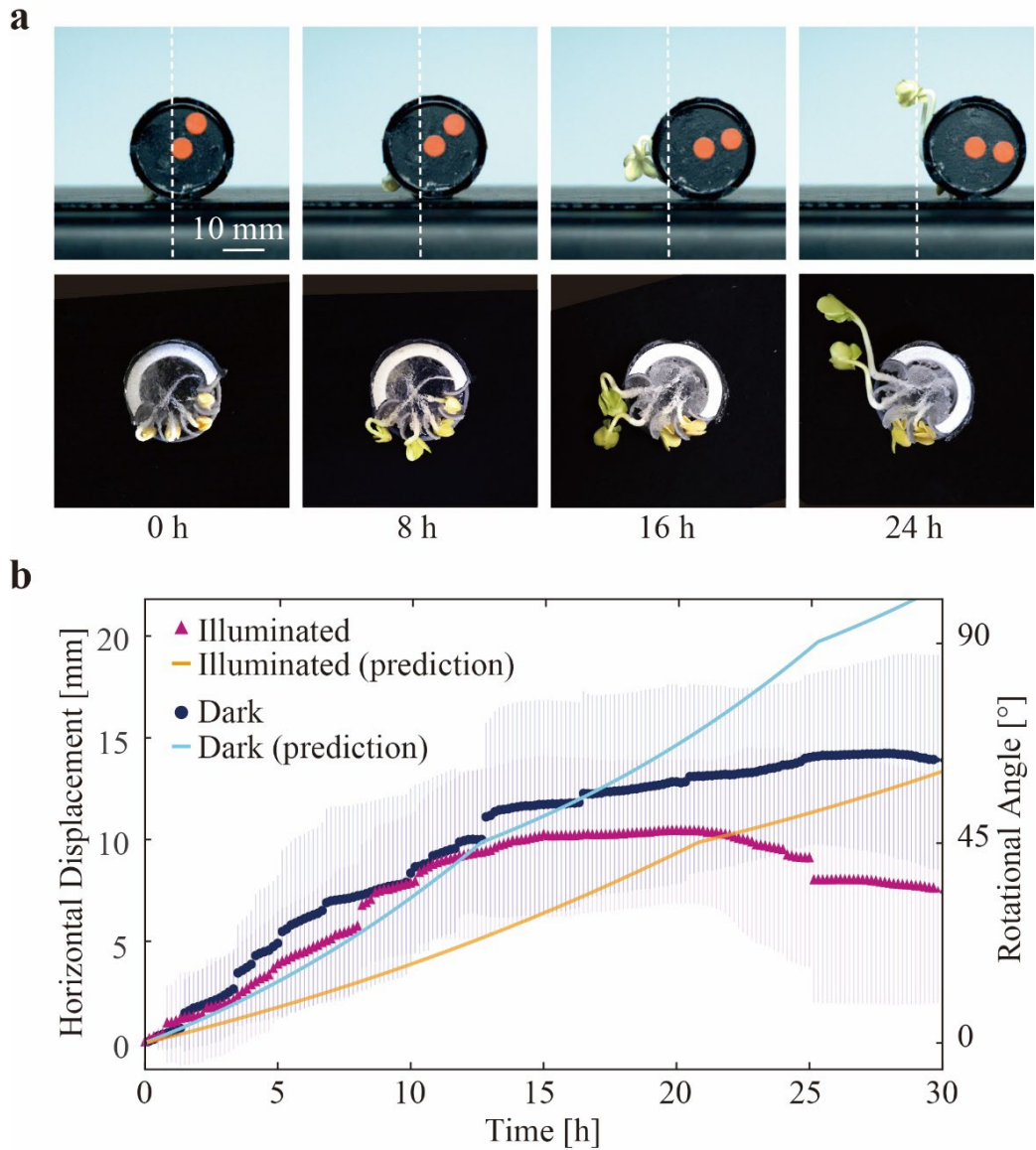

**Figure 3.** (a) Sequence of locomotion of the mobile plant robot. (b) Measured and predicted values for the horizontal movement and rotation angle of the robot as a function of the elapsed time.

## 2.3. Robotic gripper based on plant growth

Grasping is an important ability for robots working in the natural environment to perform object transportation, sampling, and perching. In order to demonstrate grasping ability powered by the plant, a robotic gripper was developed (**Figure 4**). The gripper included components such as LEDs, radish sprouts, a frame, and hydrogel that filled the housing, as illustrated in Figure 4a. In the gripper, the plants act as fingers to hold and release an object. To enable this grasping functionality, we exploited phototropism, a phenomenon of plants growing towards light.[36] As depicted in Figure 4b, when irradiated continuously with directional LED light, the hypocotyl bent approximately 90° as it grew. This behavior indicated a high degree of control over growth direction in radish sprouts in response to light exposure, enabling grasping functionality. As shown in Figure 4c(i), when the internal LED was turned on, radish sprouts started growing towards the gripper's center, demonstrating a finger-closing motion. When the outer LEDs were switched on (Figure 4c(ii)), radish sprouts grew outward, corresponding to opening fingers. Based on this principle, a pick-and-place demonstration of the gripper was performed, as shown in Figure 4d. The gripper successfully grabbed an object (a sponge with a mass of 0.1 g) and then released it. LED light solely controlled the gripper, causing the radish sprouts to automatically adapt their shape during the lifting process. This can be attributed to the flexibility of the plants, which suggests that complex control is not required for plant robots of the same type.

An additional experiment revealed that the maximum holding force of the gripper was $9.8 \pm 1.6$ mN (see the Experimental Section for more details). Notably, this error is significantly lower than that observed in the force measurement of a single plant (Figure 1e). This suggests that when multiple plants are operated in parallel, individual differences among these plants are averaged out, leading to a reduced output error. Hence, adopting such a parallel actuation approach at the system level can effectively minimize errors in the output force.

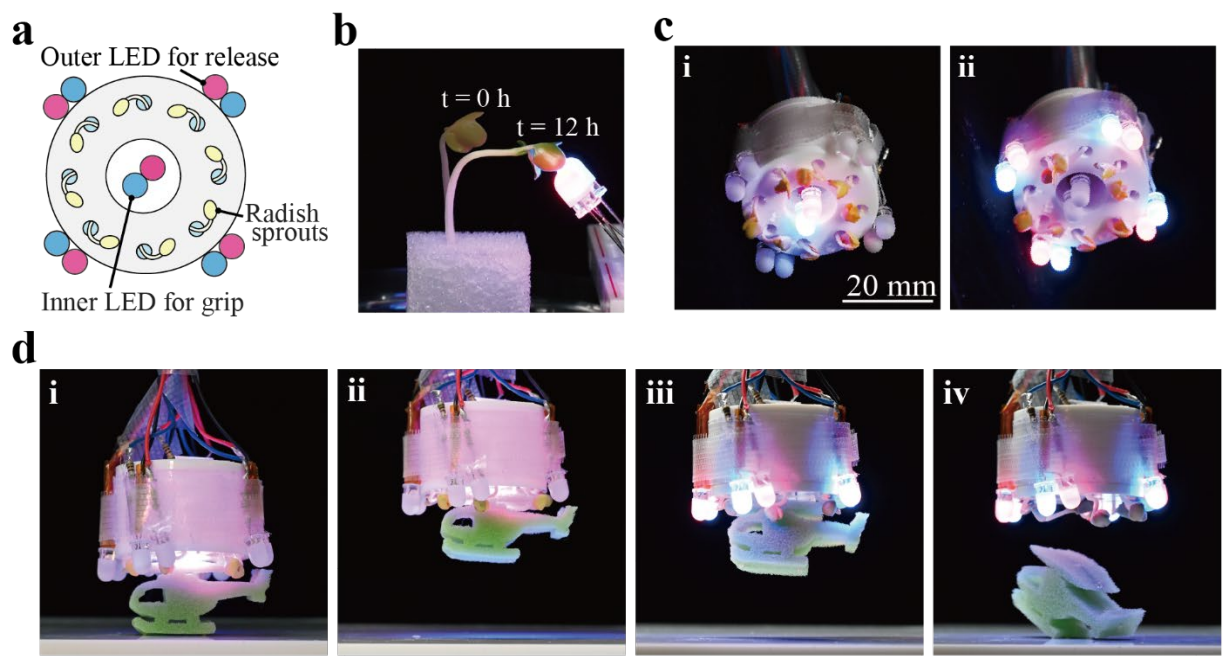

**Figure 4.** (a) Structure of the gripper. (b) Radish sprouts irradiated with LED light. (c) LED lighting arrangement for each operation. (i) Gripping by activating inner LEDs. (ii) Releasing by activating outer LEDs. (d) Pick-and-place demonstration of the gripper. (i) Grasping the object by inner LED illumination (0–11 h). (ii) Lifting the object. (iii) Outer LED illumination (11–21 h). (iv) Releasing the object.

## 3. Conclusion

In this study, we proposed and demonstrated the use of plants as actuators with integrated power sources for biodegradable robots. We revealed the actuation characteristics of radish sprouts as a target plant during the growth process and developed two types of robots: a mobile robot and a robotic gripper. The plants exhibited a maximum displacement of 76 mm within 55 h, a growth speed of 2.1 mm/h (at a length of 52.8 mm), a force of 97.5 mN, a power density of $181 \times 10^{-6}$ W/kg, and an energy density of 26.3 J/kg. This performance relied solely on water and the intrinsic biological functions of the radish sprouts, without necessitating any external power sources. However, external light, which serves as an input, is still essential, particularly for the gripper as it leverages the phenomenon of phototropism. The developed robots validated the concept of plant-powered robotic systems by displaying ground locomotion and object manipulation. A good agreement between experimental and model values was observed in the specific data of mobile robots, suggesting that obtaining the actuation characteristics of plants can enable the design and prediction of behavior in plant robots.

The plant robots presented herein display unique and intriguing characteristics. These robots operate over durations exceeding 10 h. While this may seem slow based on human standards, the relatively slow and steady movements of the robots allow them to perform tasks discreetly, without drawing the attention of potential predators in their natural environment. Moreover, the slowness of the plants is expected to enable social relationships in long-term human-robot interactions, where the social uncanniness of robotic companions could also be reduced, as discussed in a study reported quite recently.[20] Although the forces generated by the plants considered in this study may appear small, they are adequate for handling sensitive objects. Furthermore, when required, larger forces can be achieved through the simultaneous operations of multiple plants. Beyond these characteristics, the plants considered in this study are biodegradable and rely on natural sources for growth, rendering plant robots a new class of sustainable and environmentally friendly robots.

Future studies will focus on exploring the actuation characteristics of various plant species under varying environmental conditions and their integration into robotic systems. These investigations will also focus on establishing more accurate modeling and design methods.

**Experimental Section**

*Plant materials and growth chamber environment*:

Radish seeds were purchased from a supplier and sown in a hydrogel medium (SkyGel, Mebiol), where they were allowed to germinate in the dark at 25 °C for 2 days. After germination, the seed coats were removed, and the seedlings were transferred to a urethane medium for further growth and assessments. The seedlings were then placed in a growth chamber (FH-400, HiPoint) set to a relative humidity of 90% and a temperature of 25 °C. A nutrient solution with a pH of 6.5 and an electrical conductivity (EC) of 1.5 mS/cm was continuously supplied. An array of LEDs installed on the chamber ceiling provided illumination. In this environment, the photosynthetic photon flux densities (PPFDs) of red and blue light were both set to 50 $\mu mol/m^2/s$. The lights were continuously kept on for 24 h. A luminometer (SPECTROMASTER C-7000, SEKONIC) measured the PPFDs. Experimental parameters such as the temperature, pH, EC, and PPFD were set within their respective ranges used in other hydroponic culture studies.[37–42] The relative humidity was maintained at a high value of 90% to enhance plant growth.[43–46] All experiments conducted within the growth chamber were performed under the abovementioned identical environmental conditions.

*Growth displacement measurement*:

Within the growth chamber, the growth patterns of the radish sprouts in both the dark and illuminated environments were captured at 10 min intervals using a CCD camera (STC-MCS500U3V, Omron Sentech). All photographs used for measurement were taken with this camera, unless otherwise noted. Photography in the dark was done by turning on an LED solely for the duration of the shoot. The captured images underwent distortion correction processing using OpenCV. The measurement targets were the segments extending from the base of each cotyledon to the root, and their lengths were measured by image processing using ImageJ. This measurement method, calculating displacements from image pixels, featured an accuracy of $0.12 \pm 0.02$ mm/pixel. The number of samples for both environments was 15.

*Growth force measurement*:

Growth force measurements in the growth chamber were performed using a load cell (FS1M-1NB, THK) and a multimeter (2100/100, Keithley), both equipped with a transparent acrylic probe. The signal from the multimeter was recorded on a computer at 10-min intervals. After the radish contacted the load cell, the experiment continued until the generated force decreased. The standard measurement height was defined as the distance from the top of the urethane sponge to the bottom of the probe. The number of samples for each experimental item ranged from 7 to 11.

*Calculation of power density and energy density:*

After germination, the radish sprouts were washed with tap water. Furthermore, 100 seeds were measured on an electronic balance and the average value was calculated. The mass of each seed was 0.033 g. This value was used to calculate the power and energy densities of the plant. The measured force at a height of 0 mm and the velocity at the initial stem length derived from the model (4.68 mm/h in the dark environment and 2.60 mm/h in the illuminated environment) were used to calculate the power (= force × velocity). The power value was divided by the seed mass. To calculate the energy density, an average force was calculated from the measured force at heights of 0 mm and 10 mm, which was multiplied by the height difference to obtain energy (= force × distance). The resulting value was then divided by the seed mass.

*Fabrication of mobile robot*:

The robot skeleton was created using a stereolithography 3D printer (Form3, Formlabs). A 3 mm-thick styrene board was attached to both sides to prevent the robot from rolling over. Additionally, a 2 mm-thick acrylic plate, cut to an outer diameter of 25 mm and an inner diameter of 23 mm, was attached to both sides to facilitate the robot's rolling. Furthermore, the placement of the hydrogel was adjusted with a cut styrene board to centralize the robot's center of gravity. A laser cutter (Speedy 300, Trotec) was used to process these materials. Moreover, light-shielding tape was attached to the entire robot to prevent light penetration. The initial mass of the robot, including hydrogel and radish sprouts, was 5 g.

*Mobile robot movement measurement*:

The same experimental setup used in the growth displacement measurement was used for this analysis. The robot's rotation angle was calculated based on the coordinates of two positions, one at the center and another 8 mm away from the center, where red stickers were placed. The coordinates were obtained using image processing software (PFA, Photoron). The horizontal movement distance was calculated based on the rotation angle. In the dark environment, there were 11 samples, compared to 10 in the illuminated environment.

*Rolling resistance of mobile robot*:

The rolling resistance is expressed by the following formula, where $\mu_R$ is the rolling resistance coefficient and $N$ is the vertical force that the robot receives:

$$F_r = \mu_R N \qquad (2)$$

The initial mass of the robot was 5 g, and on a dry road surface, $\mu_R$ was 0.01[47]; therefore, the rolling resistance was approximately 5 mN. Since radish sprouts grew over several hours, inertial rotation did not occur, leading to a direct conversion of the growth displacement into rotational movement. In this case, the predicted rotation angle of the robot is expressed by the following formula, using the formula given in Equation (2), with the angle at which each radish sprout begins to contribute to rotation set as the initial angle (0°):

$$\theta(t) = \frac{L(t) - L_s}{R} \quad (3)$$

Here, $R$ is the radius of the robot, and $L_s$ is the length of the hypocotyl when it contacts the ground. When $\theta(t)$ reaches 45°, the next radish sprout begins to contribute to the rotation. The predicted distance of horizontal movement is calculated using the following equation:

$$d(t) = R\theta(t) \quad (4)$$

*Holding force measurement of the gripper*:

In the gripper, two types of LEDs were used: 1) a red LED with a wavelength of 623 nm, a viewing angle of 15º, and an output power of 40 mW and 2) a blue LED with a wavelength of 468 nm, a viewing angle of 15º, and an output power of 72 mW. Both LEDs were equipped with a diffusion cap. Figure S1a,b illustrates the measurement process for the maximum holding force of the gripper (denoting the maximum mass that the gripper can grasp). The sample object for grasping was constructed from two styrene board sections: one with a diameter of 20 mm and a thickness of 3 mm and the other with a diameter of 10 mm and a thickness of 15 mm. The total mass of the sample was 0.46 ± 0.06 g (N = 5). Following fabrication, the sample object and the gripper were mounted in close proximity onto a 3D-printed jig, (at a separation distance of approximately 1 mm). The jig holding both the gripper and sample object was then placed in the growth chamber, which was operating in the dark environment. The inner LEDs of the gripper were activated and left on for 18 h to grip the sample object. Thereafter, the jig was removed from the growth chamber, and the inner LEDs of the gripper were deactivated. The gripper holding the sample object was carefully detached from the jig and mounted onto a motorized linear stage (X-LRT0250BL-C-KX14N, Zaber). Subsequently, a thin rubber band was used to attach the sample object to the load cell (FS1M-1NB, THK). The linear stage elevated the gripper at a rate of 1 mm/s, and the tensile force acting on the sample object was measured through the load cell using a multimeter (2100/100, Keithley) at a sampling frequency of 30 Hz. Figure S1c displays an example of the measured force data. The maximum holding force of the gripper was computed as the sum of the mass of the sample object and the maximum tensile force corresponding to the moment when the sample object detached from the gripper. Overall, five grippers were tested, and their average holding force was reported.

**Acknowledgements**

This work was supported by the JST Fusion-Oriented Research for Disruptive Science and Technology (grant number JPMJFR2126).

**Conflict of Interest**

The authors declare no conflict of interest.

**Data Availability Statement**

The data supporting the findings of this study is available from the corresponding author upon reasonable request.